\documentclass{article}

\usepackage{arxiv}
\usepackage{latexsym}
\usepackage{amsmath}
\usepackage[caption=false,font=footnotesize]{subfig}
\usepackage[dvipdfmx]{graphicx}
\usepackage{here}
\usepackage{natbib}

\def\Underline{\setbox0\hbox\bgroup\let\\\endUnderline}
\def\endUnderline{\vphantom{y}\egroup\smash{\underline{\box0}}\\}

\graphicspath{{./fig/}}

\usepackage[varg]{txfonts}

\begin{document}

\title{Feature Representation Analysis of Deep Convolutional Neural Network using Two-stage Feature Transfer\\
-An Application for Diffuse Lung Disease Classification-}

\author{Aiga SUZUKI\\
National Institute of Advanced Industrial Science and Technology\\
1--1--1 Umezono, Tsukuba, Ibaraki, 305--8560, Japan\\
\texttt{ai-suzuki@aist.go.jp}
\And
Hidenori SAKANASHI\\
National Institute of Advanced Industrial Science and Technology\\
1--1--1 Umezono, Tsukuba, Ibaraki, 305--8560, Japan\\
\texttt{h.sakanashi@aist.go.jp}
\And
Shoji KIDO\\
Yamaguchi University\\
1677--1 Yoshida, Ube, Yamaguchi, 755--8505, Japan\\
\texttt{ai.kido@u-yamaguchi.ac.jp}
\And
Hayaru SHOUNO\\
University of Electro-Communications\\
1--5--1, Chofugaoka, Chofu, Tokyo, 182--8585, Japan\\
\texttt{shono@uec.ac.jp}
}

%

\maketitle

\begin{abstract}
Transfer learning is a machine learning technique designed to improve generalization performance by using pre-trained parameters obtained from other learning tasks.
For image recognition tasks, many previous studies have reported that, when transfer learning is applied to deep neural networks, performance improves, despite having limited training data.
This paper proposes a two-stage feature transfer learning method focusing on the recognition of textural medical images.
During the proposed method, a model is successively trained with massive amounts of natural images, some textural images, and the target images.
We applied this method to the classification task of textural X-ray computed tomography images of diffuse lung diseases.
In our experiment, the two-stage feature transfer achieves the best performance compared to a from-scratch learning and a conventional single-stage feature transfer.
We also investigated the robustness of the target dataset, based on size.
Two-stage feature transfer shows better robustness than the other two learning methods.
Moreover, we analyzed the feature representations obtained from DLDs imagery inputs for each feature transfer models using a visualization method.
We showed that the two-stage feature transfer obtains both edge and textural features of DLDs, which does not occur in conventional single-stage feature transfer models.
\end{abstract}

\section{Introduction}
In the field of computer vision and image recognition, deep convolutional neural networks (DCNNs) have been the primary model, owing to AlexNet \cite{krizhevsky2012imagenet} having had great success during the ImageNet competitions in 2012.
DCNNs are thus becoming the de facto solution for image recognition tasks.
The DCNN is a multi-layered neural network that has the same architecture as Neocognitron \cite{fukushima1980neocognitron,fukushima1988neocognitron}, inspired by biological human visual systems.
The brain's vision center has a hierarchical mechanism that understands visual stimulus \cite{hubel1962receptive}.
The DCNN uses a similar hierarchical structure to extract features by using stacks of ``convolution'' and ``spatial pooling'' operations.
The distinctive feature of a DCNN is its automation of obtaining tack feature representations, which suits the given tasks. 
Whereas DCNNs provide significant performance with image recognition tasks, they require massive amounts of training data compared to conventional machine learning models. The deep network structure exhibits higher expressive power than shallow models, which have the same complexity \cite{ba2014deep}. Alternatively, most deep models have a large number of free parameters. 
Han et al. reported that deep neural networks require one-tenth of the number of free parameters training data needed to obtain the good generalization ability \cite{han2015learning}.
However, when the acquisition of a training dataset is difficult (e.g., medical imagery), the data will sometimes be insufficient.
Generally, for learning approaches, the amount of training data has a strong effect on model performance.
Deficient training data sometimes causes generalization problems such as overfittings.

A conventional approach for overcoming data deficiency is transfer learning \cite{pan2010survey}. This is a learning technique that reutilizes knowledge gained from other learning tasks, called the ``\textit{source domain},'' to improve model performance in the desired task, called the ``\textit{target domain}.'' 
In the case of transfer learning for an image classification task, the model will first be trained to classify the source domain.
Then, it will be trained for the target domain.
In the case of DCNNs, we expect feature extraction to be improved by reutilizing its feature extraction capability.
Note that this paper distinguishes two common styles of transfer learning.
One is ``fine-tuning,'' which retrains only the classification part while maintaining the feature extraction part.
In other words, the fine-tuning style assumes that the feature extraction part
has enough ability to represent input signals.
Another is ``feature transfer,'' which retrains the entire DCNN, containing the feature extraction layers, to adopt the feature extraction part for target task.
This paper focuses on the latter case of transfer learning. In most transfer learning approaches for image recognition tasks, massive natural image datasets, such as ImageNet \cite{deng2009imagenet}, are used as the source domain \cite{tajbakhsh2016convolutional}.
The reason a natural image dataset is usually adopted is that of the availability of pre-trained models and their known performance.
However, the appropriateness of utilizing a natural image dataset when the target domain greatly differs from the natural images is slightly questionable, because features of the source domain do not appear in the target domain. 
Azizpour et al. suggested that the possibility of knowledge transfer is affected by similarities between the source and target domains.
They reported that it is preferable that transfer learning takes in similar data \cite{azizpour2016factors}.
However, only a few studies have focused on model performance variation 
by changing source and target domains, and their scope of tasks was limited to
object recognition.

This paper proposes a two-stage feature transfer method that focuses on textural image recognition.
By this method, the DCNN will successively be trained with natural and textural images as an initial state.
Afterward, all of the DCNN, which includes not only classification part but also feature extraction part, will be trained again with the textural target domain.
We will show that this type of successive and multi-domain feature transfer improves the generalization performance of the model and provides robustness with a decrease in the size of the training dataset.
Moreover, we discuss the why feature transfer on DCNNs works so well.
We visualize how feature representations of DCNNs come from different feature transfer processes and reveal that feature transfer improves feature representations of DCNNs, corresponding to both source domains.

In our experiment, we apply two-stage feature transfer to a classification task of textural X-ray high resolution computed tomography (HRCT) images of diffuse lung diseases (DLDs) and show performance improvements.

\section{Related Works and Contributions}

\cite{shouno2015transfer,gao2016holistic} applied a feature transfer to the classification of DLDs and used conventional single-staged feature transfer, which uses a natural image dataset. They reported that feature tranfer improves the classification performance over learning from scratch.
However, the appropriateness of the source domain was not discussed, despite noting that the targets were textural. \cite{christodoulidis2017multi} proposed an ensemble method that used multiple models trained with different domains for lung disorder classification. The term, ``transfer learning,'' references fine-tuning. The essence of this method entails ensemble modeling, rather than an actual transfer process.
A notable study of transfer learning in the field of medical image analysis, \cite{tajbakhsh2016convolutional}, systematically surveyed and analyzed the effects of transfer learning for various types of medical images, including textural images.
They compared transfer learning from natural images and several modern parameter initialization methods in various medical image classification tasks, which had limited amounts of training data.
They concluded that transfer learning from natural images to medical images is possible and meaningful, despite the large difference between the source and target domains. Nonetheless, the reason transfer learning works in DCNNs is still not fully understood.

In this paper, we study two-stage feature transfer, focusing on diffuse lung disease classification, making the following contributions.

\begin{figure*}[tb]
\centering
\includegraphics[width=6in]{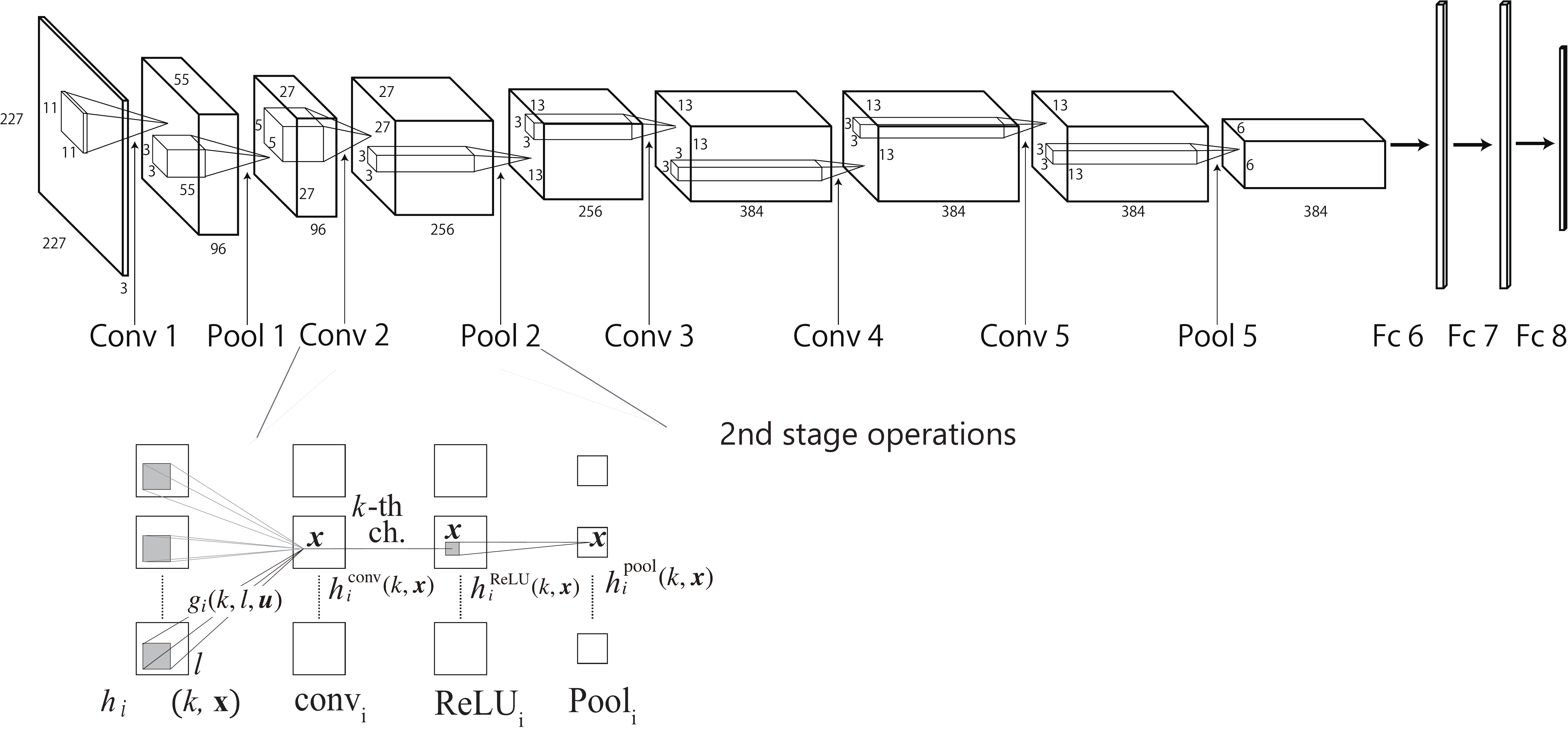}
\caption{Top: Schematic diagram of our DCNN, the same as \cite{krizhevsky2012imagenet}, or \textit{AlexNet}. 
Bottom: Details of the feature map construction. The DCNN acquires feature representation by repeating convolution and spatial pooling.}
\label{fig:alex}
\end{figure*}

\begin{itemize}
\item We demonstrate the superiority of feature transfer over fine-tuning by comparing the model performance under the same source domains.

\item We demonstrate how the source domain of feature transfer affects the performance of DCNNs by comparing learning-from-scratch, single-stage feature transfer, and our proposed method.

\item We show that transfer learning provides robust performance with a decrease in the size of the training dataset.

\item We analyze how feature representations in intermediate DCNN layers of change corresponding to the transfer processes of the feature visualization method. This change implies a DCNN mechanism of feature transfer that has not been fully researched.
\end{itemize}

\section{Deep Convolutional Neural Networks (DCNNs)}

DCNNs are well-known deep learning models, which are a type of multi-layered neural network, widely used in computer vision. The most common DCNNs consist of ``convolutions'' and ``spatial pooling'' layers, which serve as feature extractors, and fully-connected layers, which serve as classifiers. The set of convolution and pooling layers are defined as ``stages,'' in the same manner described by \cite{fukushima1988neocognitron}. The stages deform the input pattern into an intermediate representation, serving as a feature-extractor. Generally, DCNNs, which have several input channels, take 2D images and repeatedly transform them into feature maps via a stack of stages. Fig. 1 shows a schematic diagram of a typical DCNN.

To understand the feature extraction of DCNNs, let us consider the activation of $i$-th stage. Here, we denote $h_{i}(l, \mathbf{x})$ as an $l$-th channel activation, at the location, $\mathbf{x}$, in the $i$-th stage. Convolution layers provide convolutional filtering to derive feature maps (i.e., activations) from previous stages. The activation of the convolution layer is written as
\begin{equation}
h_{i}^{\text{conv}}(k, \mathbf{x})
= \sum_{l, \mathbf{u}}
g_i (k, l, \mathbf{u})~h_{i-1}(l, \mathbf{x} - \mathbf{u}),
\label{eq:conv}
\end{equation}
where $k$ is the channel of the derived feature map, and $g_i (k, l, \mathbf{u})$ is the convolution kernel (i.e., a ``\textit{filter tensor}''). Eq. \ref{eq:conv} shows that the convolution layer makes a feature map as an inner product of a filter tensor, $g_i$, and all regions of input. Most neural networks modulate responses of each layer with an activation function to provide a non-linearity. We chose the rectified linear unit (ReLU), commonly used in deep neural networks, as the activation function. Following the convolution layer, all feature maps, $h_{i}(k, \mathbf{x})$, are modulated with ReLU.
\begin{equation}
h_{i}^{\text{relu}}(k, \mathbf{x}) = \max
\left(0, h_{i}^{\text{conv}}(k, \mathbf{x})\right)
\end{equation}

The pooling layer gathers spatial neighbors to reduce the repercussions of local pattern deformations and the dimensionality of the feature map. The response to the pooling layer of the feature map, $h_{i}(l, x)$, is computed as
\begin{equation}
  h_{i}^{\text{pool}}(k, \mathbf{x}) = \max_{\mathbf{r} \in N(\mathbf{x}))}
  \left(0, h_{i}(k, \mathbf{r})\right),
\end{equation}
where $N(\mathbf{x})$ is the spatial neighbor at location, $x$, in the feature map. This type of pooling operation, which uses the maximum value of spatial neighbors as a representative value, is called ``max-pooling.''

These layers appear as early DCNN layers, which sum inputs and provide well-posed inputs for a given task. Trainable parameters of these formulations are the filter tensors, $g_i$.

The latter layers of DCNNs (i.e. ``Fc $n$,'' in Fig. \ref{fig:alex}) are fully-connected layers. In Fig. \ref{fig:alex}, extracted feature representations of the input image appear as the first fully-connected layer, ``Fc 6.'' Layers, ``Fc7'' and ``Fc8'' comprise a multi-layered perceptron, which plays the role of classifier.

The most remarkable trait of DCNNs is its effective feature representation, corresponding to tasks that are obtained as an intermediate representation of the feature extraction parts, consisting of convolution and spatial-pooling layers. These are obtained via a back-propagation algorithm, which minimizes classification errors.

\begin{figure}[tb]
\centering
\includegraphics[width=4.1in]{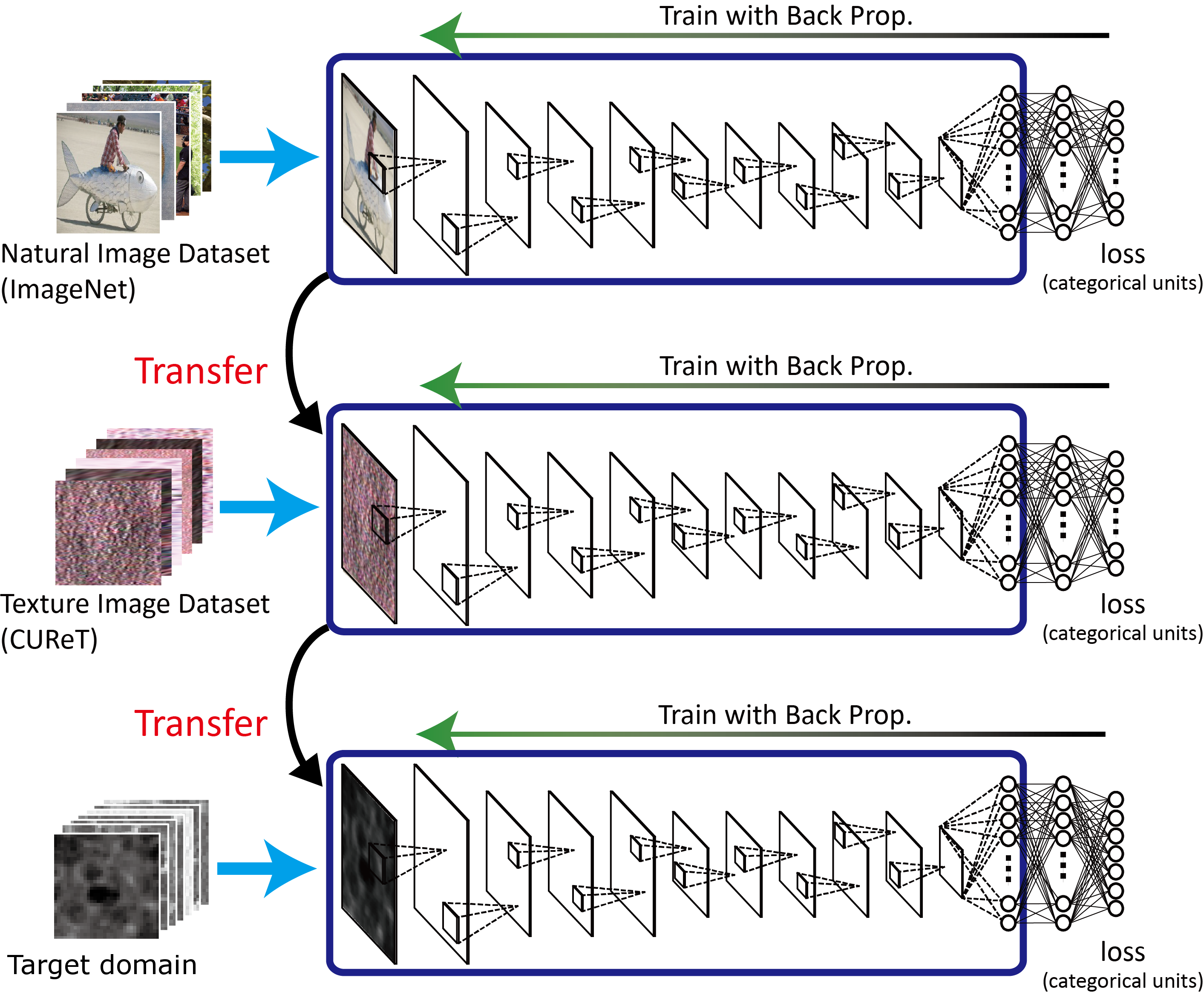}
\caption{
Schematic diagram of two-stage feature transfer for analyzing DLD HRCT patterns. The DCNN is first trained with natural images to obtain a good feature representation as the initial state. Afterward, it transfers to the more effective domain (i.e., texture dataset) to obtain the feature representation suited for texture-like patterns. Then, finally it trains with the target domain.}
\label{2stl}
\end{figure}

\begin{figure*}[tb]
\centering
\includegraphics[width=4.5in]{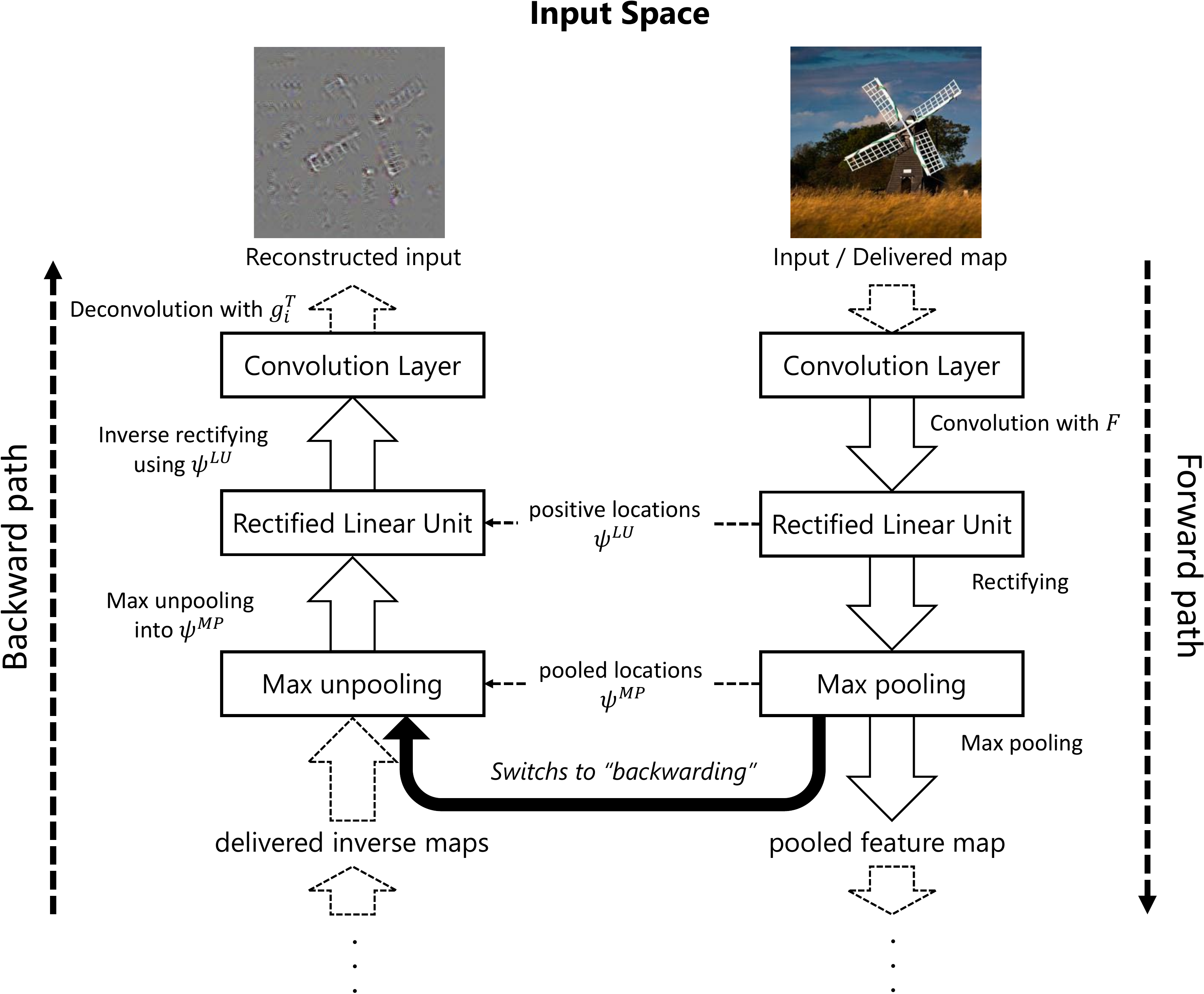}
\caption{
A feature visualization flow using DeSaliNet.
The feature map to visualize
is calculated at a forward propagation (right).
When visualizing neuronal activations,
the feature map is switched to backward visualization path (left),
which consists of inverse maps of each forward layers,
and is backpropagated into input space as a saliency image.
}
\label{fig:desali}
\end{figure*}

\section{Methods}

\subsection{Two-Stage Feature Transfer}
Transfer learning is a technique that reutilizes feature expressions that come from similar tasks \cite{pan2010survey}. This paper proposes a two-stage feature transfer method focused on textural recognition tasks.

Fig. \ref{2stl} shows the schematic diagram of two-stage feature transfer, which, for DCNNs, means the reutilization of the feature extraction parts of the pre-trained network.
These parts consist of convolution layers and no classification layers.
Thus, the fully-connected layers (i.e., ``Fc7'' and ``Fc8'') are cut off from their connections, as shown in Fig. \ref{fig:alex}.
After reconfiguring the network, we randomly initialize connections of the classifier part
\footnote{Fully-connected weights, without a softmax layer (e.g., ``Fc8'') can be reused as the initial state for the transfer. In our experiment, however, the resulting performance has been worsened.}
and train the entire DCNN again using back-propagation. Thus, feature transfer utilizes the feature extraction parts from other domains as its initial state.

In our proposed method, we first train the DCNN with massive natural images in the same manner as conventional feature transfer. At this stage, we expect that all connections are well-trained for extracting visual features from input images of natural scenes, such as edge structures \cite{krizhevsky2012imagenet,zeiler2014visualizing}. Second, we apply feature transfer again, using the texture image dataset and natural images to acquire better feature representation and fitting for the textural images, which do not appear in the natural images.

\subsection{Feature Visualization}

For analysis, to understand the mechanism of knowledge transfer in DCNNs, and to reveal how feature transfer influences improvements, we should discuss what is attended by the DCNN feature extraction process.
We adopt DeSaliNet, proposed by \cite{mahendran2016salient}, as our feature visualization method.
This includes similar methods proposed by \cite{zeiler2014visualizing} and \cite{simonyan2013deep} as its special cases. DeSaliNet reveals which input component influences the feature representation of the feature extraction parts. Fig. \ref{fig:desali} shows the process flow of a feature visualization using DeSaliNet.

The main idea of DeSaliNet is to propagate the feature map backward into the input space. DeSaliNet construes DCNN operations as functions and describes itself as a composite function. Let $\phi^{(i)}$ be a map to the $i$-th layer's feature map that we want to visualize. $\phi^{(i)}$ can thus be denoted by each layer activation, up to the $i$-th layer, as
\begin{equation}
  \phi^{(i)} = h^{L_i}_i \circ \cdots \circ h^{L_1}_1,
\end{equation}
where $L_i$ is the layer type, such as convolution, max-pooling, and ReLU. 
Here, we also denote the ``backward path,'' $\phi^{(i)\dagger}$, which is illustrated on the left side of Fig. \ref{fig:desali} as an inverse map of $\phi^{(i)}$.
\begin{equation}
  \phi^{(i)\dagger}= h^{L_1\dagger}_1 \circ \cdots \circ h^{L_i\dagger}_i,
\end{equation}
where $h^{L_i\dagger}_i$ denotes inverse maps associated with its corresponding layer, $h^{L_i}_i$. Details of each inverse map are discussed in Appendix A.1. Then, the visualization result, $\phi^{(i)\dagger}(h)$, is obtained as a member of the input space.

The origin of the visualization method, based on backward propagation, is the selective attention model \cite{fukushima1987neural,shuono1995connected}. This type of feature visualization enables us to analyze \textit{what component is paid attention to} in input images, in contrast to saliency maps \cite{simonyan2013deep}, which analyze \textit{where it is paid attention to}. Textural images are ``what-based,"  because the textural images do not have locality as a characteristic.

\section{Materials}

\subsection{Target Domain}

\begin{figure}[tb]
\centering
\includegraphics[width=5.3in]{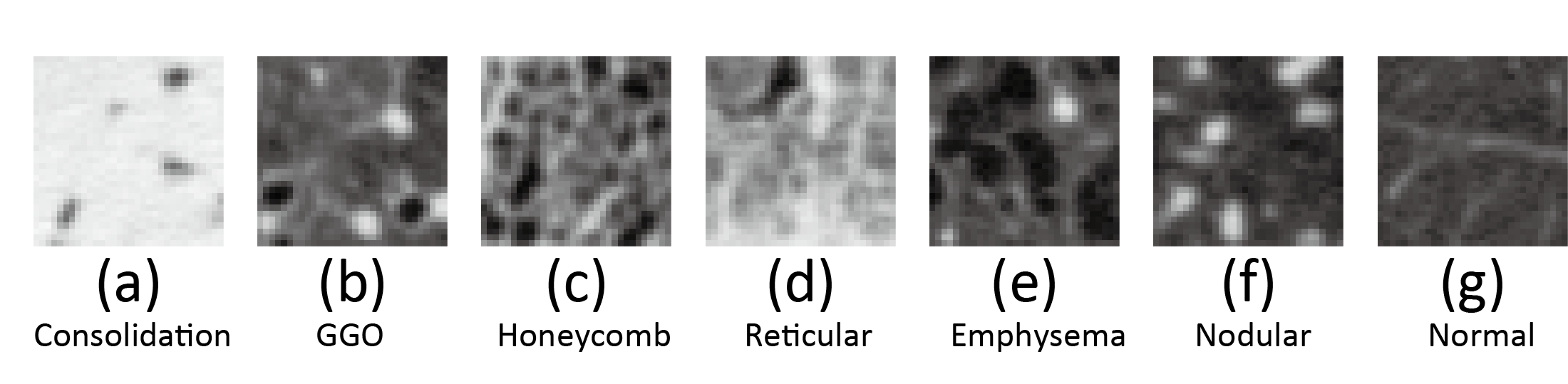}
\caption{
Typical HRCT images of diffuse lung diseases: (a) consolidations (CON); (b) ground-glass opacities (GGO); (c) honeycombing (HCM); (d) reticular opacities (RET); (e) emphysematous changes(CON); (f) nodular opacities (NOR); and (g) normal (NOR).
}
\label{fig:dlds}
\end{figure}

We examined the effectiveness of our proposed two-stage feature transfer method with the classification of X-ray and HRCT DLDs. DLD is a collective term for lung disorders that can spread to large areas of the lung. X-ray HRCT is effective for finding early-stage DLDs when they are small and mild. DLD conditions are seen as textural patterns on HRCTs. In this work, these patterns are classified into seven classes: consolidations (CON), ground-glass opacities (GGO), honeycombing (HCM), reticular opacities (RET), emphysematous changes(CON), nodular opacities (NOR), and normal (NOR). These categorizations were introduced by \cite{uchiyama2003quantitative}. Fig. \ref{fig:dlds} shows portions of HRCT images for each class.

The DLD image dataset was acquired from Osaka University Hospital, Osaka, Japan. We collected 117 HRCT scans from different subjects. Each slice was converted to gray-scale images with a resolution of 512 × 512 pixels and slice-thickness of 1.0 [mm]. Lung region slices were annotated for their seven types of patterns by experienced radiologists. The annotation region shapes and their labels were the results of diagnoses by three physicians. The annotated CT images were partitioned into regions of interest (ROI) patches, which were 32 $\times$ 32 pixels, corresponding to about 4 [cm${}^2$]. This is a small ROI size for DCNN input. Thus, we magnified them by 224 $\times$ 224 pixels using bicubic interpolation. Therefore, from these operations, we collected 169 patches for CON, 655 for GGO, 355 HCM, 276 for RET, 4702 for RET, 827 for NOD, and 5726 for NOR. We then divided these patches for DCNN training and for an evaluation, because each class does not contain patches from the same patients. For the training, we used 143 CONs, 609 GGOs, 282 HCMs, 210 RETs, 4406 EMPs, 762 NODs, and 5371 NORs. The remaining 26 CONs, 46 GGOs, 73 HCMs, 66 RETs, 296 EMPs, 65 NODs, and 355 NORs were used for the evaluation.

\subsection{Source Domains}

\begin{figure}[tb]
\centering
\includegraphics[width=3.1in]{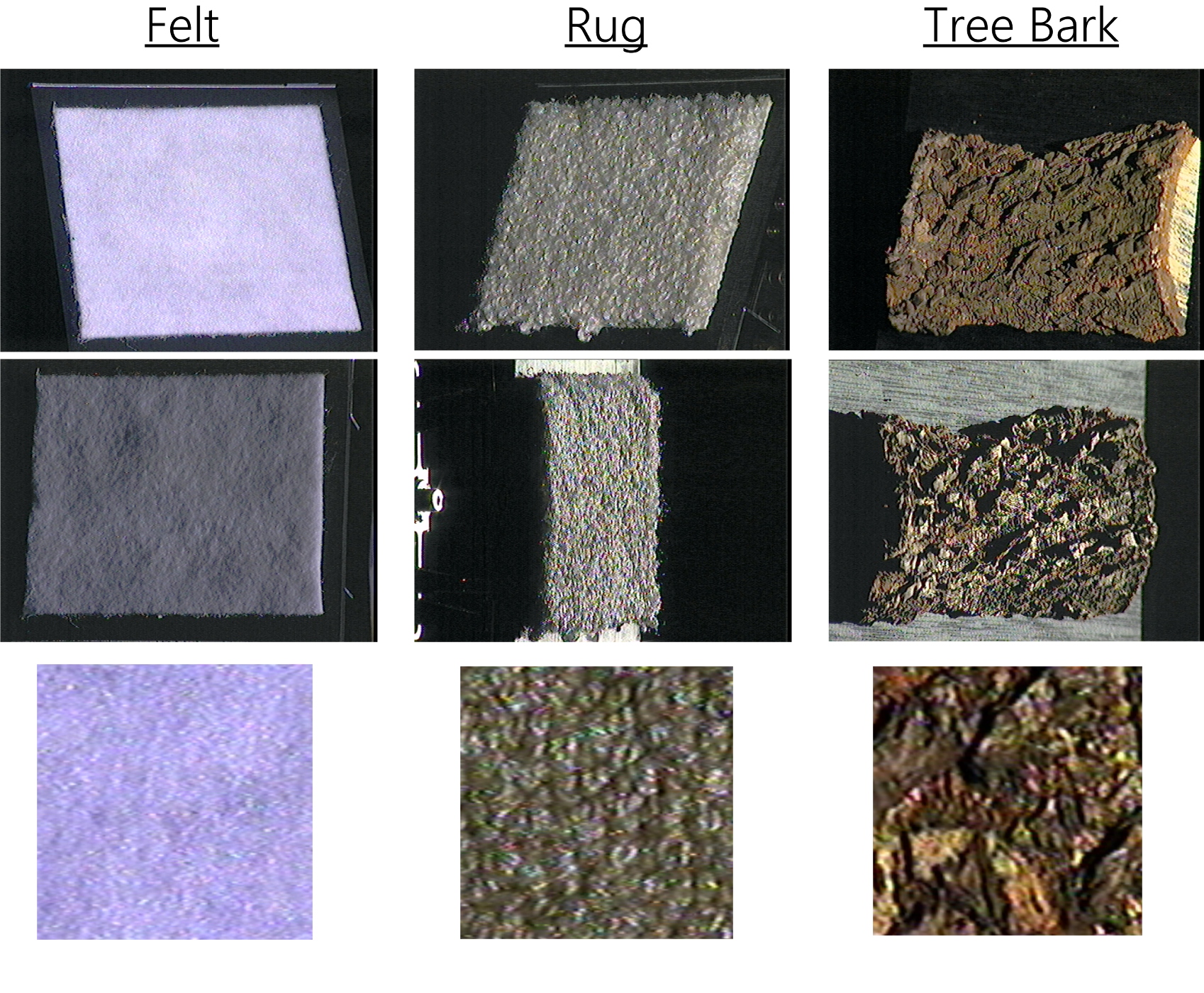}
\caption{
Examples of textural images comprising the CUReT database. Top and middle rows: entire images of ``Felt,'' ``Rug,'' and ``Tree Bark'' classes.
Bottom: cropped and resized images used as input for the DCNN.
}
\label{fig:curet}
\end{figure}

Two-stage feature transfer uses both natural image and texture datasets. We used ILSVRC2012 dataset, 
which is a subset of ImageNet \cite{deng2009imagenet}, as the natural image dataset.
We also used the Columbia-Utrecht Reflectance and Texture Database (CUReT) \cite{dana1999reflectance} as the texture dataset, as provided by Columbia University and Utrecht University. Fig. \ref{fig:curet} shows examples of textural images in CUReT database. The database contains macro photographs of 61 classes of real-world textures. Each class has approximately 200 samples. Each sample was imaged under various combinations of illumination and viewing angles. To train DCNNs, we cropped the textured regions and resized them into 224 $\times$ 224 to accommodate network input.

\section{Experiments}

The network structure used in this work is exactly same as AlexNet \cite{krizhevsky2012imagenet}, illustrated in Fig. \ref{fig:alex}.
We trained the network using momentum stochastic gradient descent with a momentum of $0.9$ and a dropout rate of $0.5$.
When the network was trained for the first time, we set the learning rate to 0.05.
Otherwise, we set the learning rate to 0.0005 because it is reported that small learning rate is preferable for pre-trained networks in
\cite{tajbakhsh2016convolutional}.
We trained the network until training loss plateaus, as to steadily converge the network parameters.

For evaluation metrics, we used accuracy, recall, precision, and F1-score.
Accuracy is the proportion of correct predictions to the total number of predictions.
Recall is the fraction of samples collectively classified over the number of samples of its class.
Precision is the fraction of samples correctly classified as class, $c$, over all samples classified as a class $c$.
Recall is an index of oversights, whereas precision is an index of over-detections.
The F1-score is a harmonic mean between precision and recall: 
$\frac { 2 \cdot Precision \cdot Recall} {Precision + Recall}$.

In our experiment, we compared models from different learning processes,
as follows.
\begin{enumerate}

\item Learning a randomly initialized model from scratch in the most naive way (i.e., no feature transfer)
\item Feature transfer from textural images, i.e., CUReT database
\item Feature transfer from natural images, i.e., ILSVRC 2012 dataset
\item Two-stage feature transfer, training the DCNN from ILSVRC 2012 and CUReT, sequentially \textit{(proposed)}

\end{enumerate}

\subsection{Classification Performance}

First, we compared the classification performance of each models (1) $\sim$ (4).
In addition to this, to reveal the effectiveness of feature transfer, we also compared to fine-tuning models as follows:
\begin{itemize}
  \item[(a)] Fine-tuning from natural images (ILSVRC 2012 dataset)
  \item[(b)] Fine-tuning from textural images (CUReT database)
\end{itemize}

Results are shown in Table \ref{table:accuracies}.
Feature transfer models (1) $\sim$ (4) surpass fine-tuning models (a) and (b) at all classification performances.
This suggests that the feature representation obtained in natural and textural images does not suit for DLD classification,
in other words, feature extraction part ought to be retrained with the target domain.
Also, the two-stage feature transfer (4) displays the best performance.
However, despite feature transfer, the model (2) using CUReT performed worse than learning from scratch.
This implies that CUReT, by itself, is useless as the source domain for conventional feature transfer.

\begin{table*}[t]
\begin{center}
\caption{Classification performance comparison for test data}
\label{table:accuracies}
  \begin{tabular}{c|c|c|c|c||c|c}
    \hline 
    & (1) & (2) & (3) & (4) & (a) & (b) \\ \hline
    Transfer & None &  \multicolumn{2}{|c|}{single-stage (conventional)} & two-stage \textit{(proposed)} & \multicolumn{2}{|c}{fine-tuning}\\ \hline
    Accuracy & 0.9277  & 0.9201 & 0.9558 & 0.9601 & 0.7735 & 0.8263 \\
    Precision & 0.9583 & 0.9412 & 0.9484 & 0.9739 & 0.7842 & 0.8345 \\
    Recall & 0.9590    & 0.9417 & 0.9471 & 0.9719 & 0.7735& 0.8263 \\
    F1-score& 0.9583   & 0.9411 & 0.9470 & 0.9724 & 0.7675 & 0.8228\\
    \hline
  \end{tabular}
\end{center}
\end{table*}

\subsection{Model Robustness for amounts of Training data}

\begin{figure}[tb]
  \begin{tabular}{c}
    \begin{minipage}{0.5\hsize}
\centering
\includegraphics[width=3.1in]{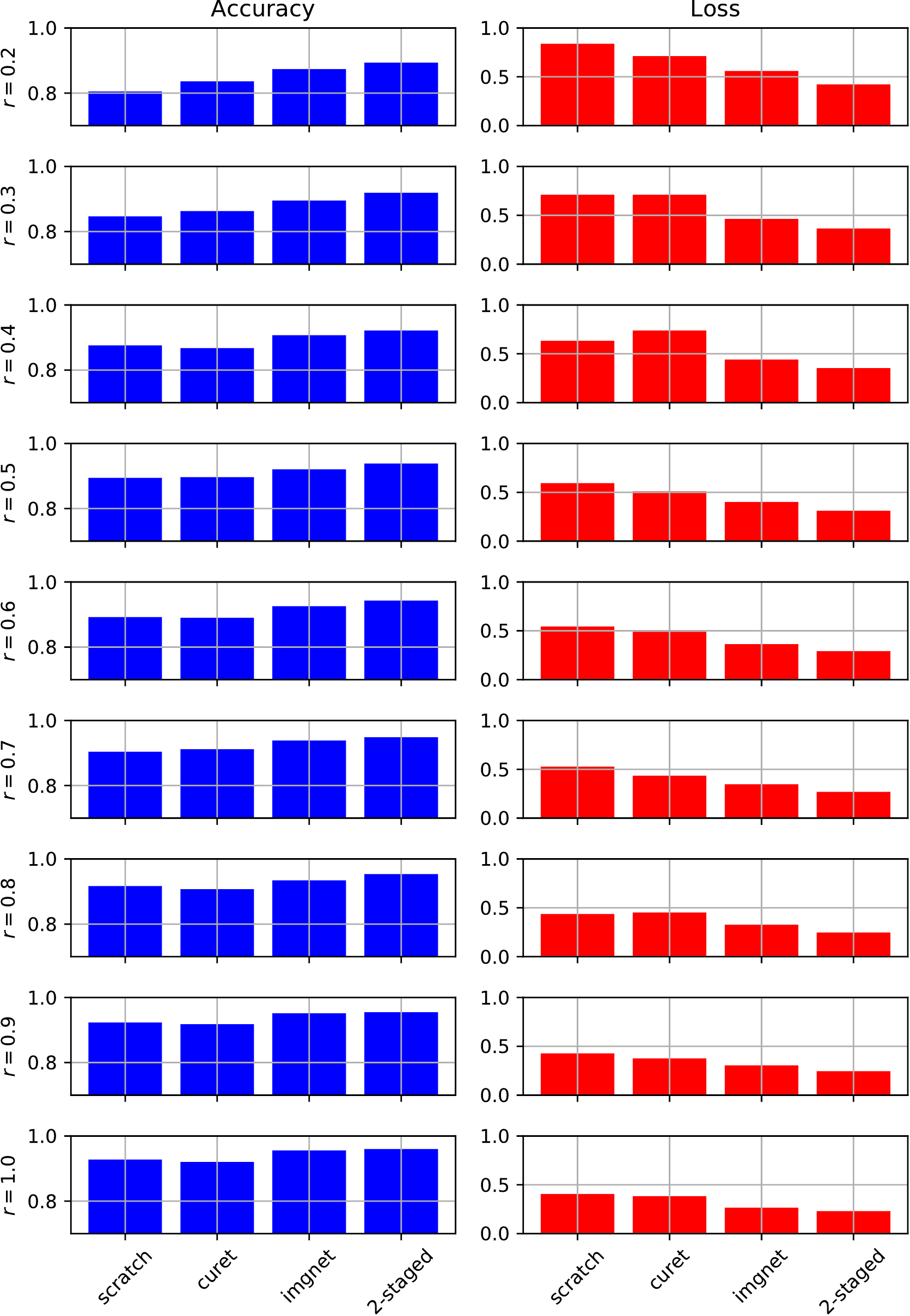}
\caption{
Performance comparisons of each amount of training data:
(Left) classification accuracies of DLDs;
  (Right) cross-entropy losses of ``Fc8'' in Fig.\ref{fig:alex}.
  Each bar, from left to right, shows the learning processes:
  (1) learning from scratch;
  (2) single-staged feature transfer with CUReT;
  (3) single-staged feature transfer with ImageNet;
  and (4) our proposed two-stage feature transfer.
}
\label{fig:comp_bar}
    \end{minipage}
    \begin{minipage}{0.5\hsize}
\centering
\includegraphics[width=3.1in]{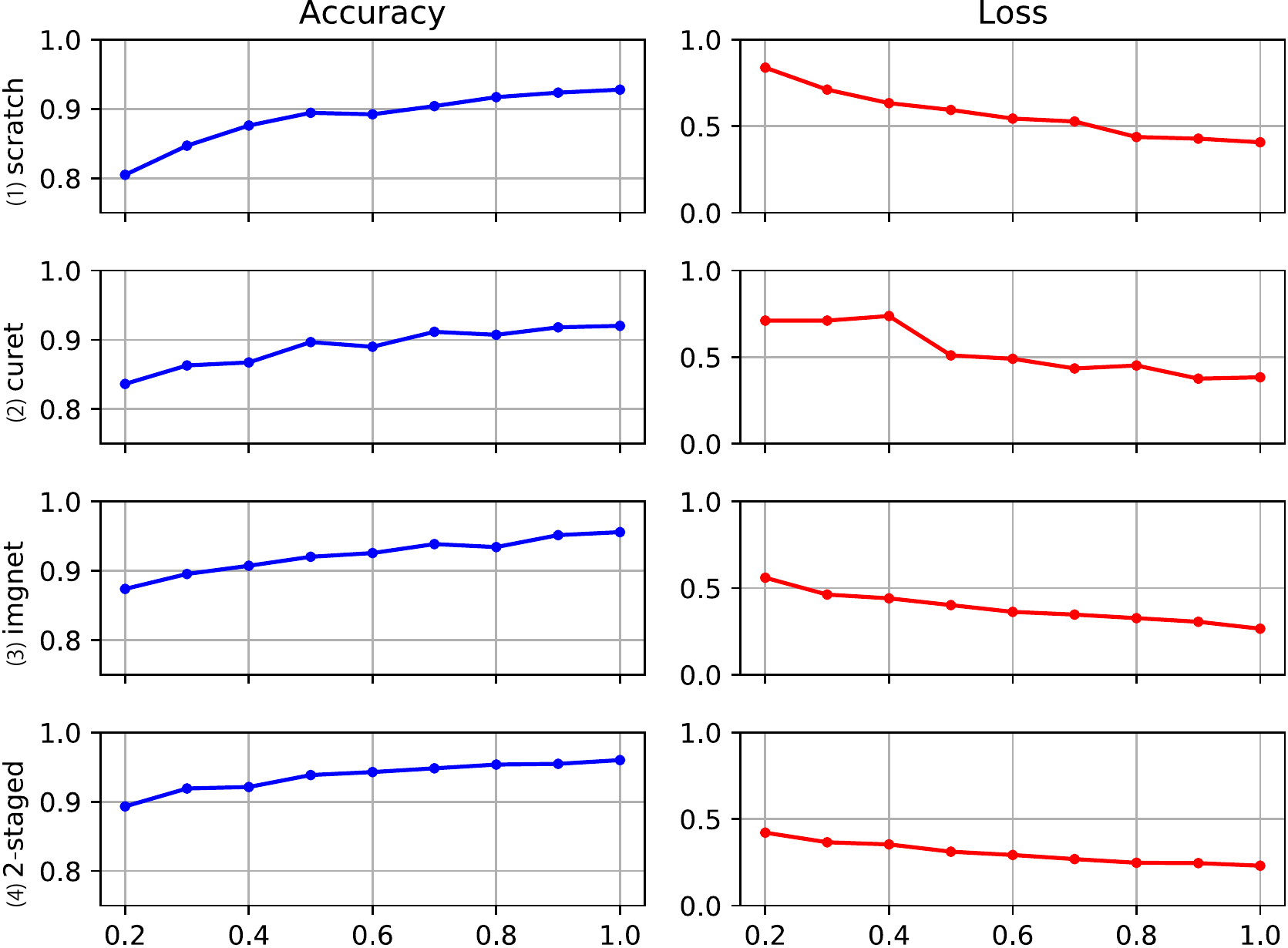}
\caption{
Fluctuation comparisons of each learning process:
(Left) classification accuracies for validation data;
  (Right) softmax losses of validation data.
  Each row (1) $\sim$ (4), from top to bottom, shows the learning processes.
}
\label{fig:comp_line}
    \end{minipage}
  \end{tabular}
\end{figure}

Moreover, we demonstrated how the robustness of each model, with respect the decrease in the amount of training data, improved.
We transitioned the accuracies and losses of the softmax layer (i.e., ``Fc8'' in Fig. \ref{fig:alex})
by changing the amount of DLD training samples by the ratio, $r$, from 20[\%] to 100[\%]
\footnote{For example, when $r=1.0$ and $r=0.5$, the amounts of training DLD examples are $927$ and $434$, respectively.
Proportions of each class are retaining.}.
Fig. \ref{fig:comp_bar} shows the models' performance comparison.
In all cases, two-stage feature transfer showed the best performance for both accuracy and loss, especially in the case of a small training dataset.


Fig. \ref{fig:comp_line} shows the fluctuation of model performance with a decline in the amount of DLDs images. To quantify the degree of model robustness, we assumed that these variations have linearity to the amount of data, and compared slopes, $A$, of the linear regression model: $\text{Accuracy} = Ar + b$, 
where $r$ is the percentage of data, and $b$ is the intercept coefficient. Clearly, a small absolute value of slope indicates that the model is more robust with $r$. All feature transfer models show better results than learning from scratch, as shown in Table \ref{table:slopes}. two-stage feature transfer showed the best robustness, both with accuracy and with loss.

\begin{table}[tb]
\begin{center}
\caption{Variations of model performances in each process}
\label{table:slopes}
  \begin{tabular}{c|c|c|c|c}
    \hline 
    & (1) & (2) & (3) & (4) \\ \hline
    Slopes of accuracies & 1.3560 & 0.9920 & 0.9475& 0.7479 \\
    Slopes of losses     & -0.5054& -0.4938& -0.3221& -0.2230\\
    \hline
  \end{tabular}
\end{center}
\end{table}

\section{Analysis of the Feature Extraction}

\begin{figure*}[tb]
\centering
\includegraphics[width=5.7in]{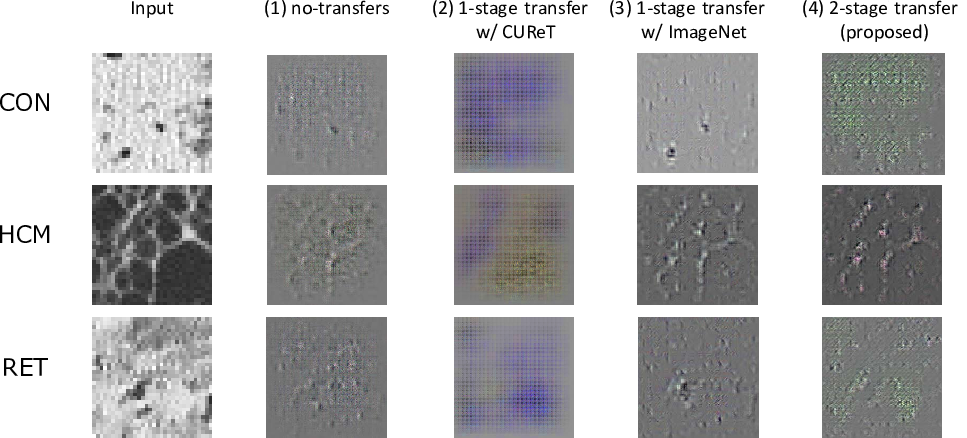}
\caption{
Visualization results of Fc 6's feature maps came from DLD images.
The leftmost figures show the DCNN inputs.
Each row represents the input DLD images,
which are CON, HCM, and RET, respectively.
Each column represents the DCNN learning processes.
Bright or colored regions indicate that the corresponding components of inputs have a strong effect on feature maps.
}
\label{fig:visualize_dlds}
\end{figure*}

Fig. \ref{fig:visualize_dlds} shows the visualization results of extracted features 
(i.e., the activation of Fc6 in Fig. \ref{fig:alex}) for each model, from (1) to (4).
Model (1), learned from scratch, did not show salient activities in any region of input.
This suggests that the model could not extract meaningful features from inputs because of the lack of training data.
Model (2), transferred from textural images, showed activation in the regions where the textural structure appeared (e.g., pits of CON or cyst wall contours of HCM).
Alternatively, model (3), transferred from natural images, showed activations in the regions where edge structures appeared
(e.g., entire of CON or bottom right of RET, which both colored in blue).
It is intuitive, considering that the models trained for natural images show an activation for edge structures (e.g., the object contours and lines), as reported by most studies on the visualization of DCNNs \cite{zeiler2014visualizing,simonyan2013deep,mahendran2016salient}.
Interestingly, Model (4), which came from two-stage feature transfer, responded to both edge and textural structures.
The models (2) and (3) show the strong responses to the edge and textural
regions respectively.
In contrast, we can see these models show weak responses to the opposite regions.
Given the results of (2) and (3), such feature representations seem to be additively obtained from both natural images and textural domains during two-stage feature transfer.
Performance improvements occur because the DCNNs obtain better feature representation, which suits textural patterns with the two-stage feature transfer.

\section{Conclusion}

We proposed a two-stage feature transfer, which improved the performance of DCNNs for classification tasks of textural images, as an extension of conventional transfer learning methods, which use a single domain as the source. We applied two-stage feature transfer to the classification of HRCT images of lung diseases and demonstrated that two-stage feature transfer improves classification performance and robustness while decreasing the amount of training data, compared to learning from scratch and conventional transfer learning. To assess these improvements, we analyzed and compared each feature representation using a feature visualization method.
Two-stage feature transfer seems to have provided appropriate feature representations for both edge and textural structures transferred from natural images and textural images, respectively. These results indicate the consequence of source domain selection.

\subsection*{Acknowledgment}
  This work is partially supported by Grant-in-Aids for Scientific Research KAKENHI (C) 16K00328, and Innovative Areas 16H01452, MEXT, Japan.
  We really appreciate Prof. Honda, Osaka University Hospital providing the HRCT images of DLDs.

\appendix
\section{DeSaliNet's Inverse Maps}

This appendix provides details of the inverse maps used in DeSaliNet \cite{mahendran2016salient}. In Eq. 5, each $\phi^{(L_i)\dagger}_i$ denotes the inverse map of each forward operation, $\phi^{(L_i)}_i$, where the $L_i$ is a layer type of the $i$-th stage. DeSaliNet considers only the case where $L_i \in \{\text{convolution, max-pooling, ReLU}\}$,
otherwise the layers be ignored. This results in an identity map.

\begin{description}
\item[Convolution layer]\mbox{}\\
Let $h_i(l, \mathbf{x})$ be a feature map of the $l$-th channel, where it is in the position, $\mathbf{x}$. The inverse map of the convolution layer, $\phi^{\text{conv}\dagger}_i$, called ``deconvolution,'' denoted as 
\begin{equation}
\phi^{\text{conv}\dagger}\left(h_i(l, \mathbf{x})\right)
= \sum_{l, \mathbf{u}}
g_i(k, i, S(\mathbf{u}))~h_{i}(l, \mathbf{x} - \mathbf{u}),
\label{eq:deconv}
\end{equation}
where
\begin{equation}
\begin{array}{cccc}
    S: & \mathbb{Z}^2 & \longrightarrow & \mathbb{Z}^2 \\
    & \rotatebox{90}{$\in$} && \rotatebox{90}{$\in$} \\
        & \left( \begin{array}{c} x\\ y \end{array} \right) & 
        \longmapsto & \left(\begin{array}{c} y\\ x  \end{array} \right).
\end{array}
\label{eq:pos_switch}
\end{equation}
Eqs. \ref{eq:deconv} and \ref{eq:pos_switch} indicate that the deconvolution layer is a convolution for feature maps having a transposed filter tensor, $\mathbf{g}_i$.

\hspace{1em}
\item[Max-pooling layer] \mbox{}\\
The inverse map of the max-pooling layer, $\phi^{\text{MP}\dagger}_i$, is denoted as
\begin{equation}
\phi^{\text{MP}\dagger}(h_i(l, \mathbf{x}))
= \left\{
    \begin{array}{ll}
      h_i(l, \mathbf{x}) & (x \in \psi^\text{MP}_i) \\
      0 & \text{otherwise}
    \end{array},
  \right.
\end{equation}
where $\psi^{MP}_i$ contains the stored maximum value locations of forward calculation in max-pooling. The pooled map is sparsely restored into a maximum position.

\hspace{1em}
\item[Rectifying layer]\mbox{}\\
The inverse map of the ReLU layer, $\phi^{\text{ReLU}\dagger}_i$, is denoted as
\begin{equation}
\phi^{\text{ReLU}\dagger}(h_i(l, \mathbf{x}))
= \left\{
    \begin{array}{ll}
      h_i(l, \mathbf{x}) & (x \in \psi^\text{LU}_i) \\
      0 & \text{otherwise},
    \end{array}
  \right.
\label{eq:unrelu}
\end{equation}
where $\psi^\text{LU}_i$ is stored positive locations in the forward ReLU caculation, where zero was not modulated.
\end{description}
\bibliographystyle{unsrt}
\bibliography{bib}

%
\end{document}